\documentclass[conference]{IEEEtran}
\IEEEoverridecommandlockouts
\usepackage{cite}
\usepackage{amsmath,amssymb,amsfonts}
\usepackage{algorithmic}
\usepackage{graphicx}
\usepackage{textcomp}
\usepackage{xcolor}
\usepackage{hyperref}
\usepackage{float}
\usepackage{flushend}
\def\BibTeX{{\rm B\kern-.05em{\sc i\kern-.025em b}\kern-.08em
    T\kern-.1667em\lower.7ex\hbox{E}\kern-.125emX}}

\usepackage{todonotes}

\newcommand{\emailaddress}[1]{\href{mailto:#1}{#1}}
\newcommand{\mathc}[1]{\mathcal{#1}}
\definecolor{d_green}{RGB}{65, 117, 5}
\definecolor{d_red}{RGB}{255, 0, 0}

\makeatletter 
\newcommand{\linebreakand}{%
  \end{@IEEEauthorhalign}
  \hfill\mbox{}\par
  \mbox{}\hfill\begin{@IEEEauthorhalign}
}
\makeatother 

\begin{document}

\title{Reinforcement learning for Energies of the future and carbon neutrality: a Challenge Design}

\author{\IEEEauthorblockN{Gaëtan Serré\textsuperscript{1}}
\IEEEauthorblockA{\emailaddress{gaetan.serre@universite-paris-saclay.fr}}
\and
\IEEEauthorblockN{ Eva Boguslawski\textsuperscript{1, 3}}
\IEEEauthorblockA{\emailaddress{eva.boguslawski@rte-france.com}}
\and
\IEEEauthorblockN{ Benjamin Donnot\textsuperscript{3}}
\IEEEauthorblockA{\emailaddress{benjamin.donnot@rte-france.com}}
\and
\linebreakand
\IEEEauthorblockN{ Adrien Pav\~ao\textsuperscript{1}}
\IEEEauthorblockA{\emailaddress{adrien.pavao@gmail.com}}
\and
\IEEEauthorblockN{ Isabelle Guyon\textsuperscript{1, 2}}
\IEEEauthorblockA{\emailaddress{guyon@chalearn.org}}
\and
\IEEEauthorblockN{ Antoine Marot\textsuperscript{3}}
\IEEEauthorblockA{\emailaddress{antoine.marot@rte-france.com}}
\and
\linebreakand
{\footnotesize \textsuperscript{1}LISN/CNRS/INRIA, University of Paris-Saclay, Gif-Sur-Yvette, France,
\footnotesize\textsuperscript{2}ChaLearn,
\footnotesize\textsuperscript{3}RTE France}
}

\maketitle

\begin{abstract}
  Current rapid changes in climate increase the urgency to change energy production and
  consumption management, in order to reduce carbon and other greenhouse gas production.
  In this context, the French electricity network management company RTE (Réseau de Transport d'Électricité)
  has recently published the results of an extensive study outlining various scenarios for tomorrow’s
  French power management~\cite{futurs_energetiques}.
  We propose a challenge that will test the viability of such scenarios~\cite{electricity_mix}.
  The goal is to control electricity transportation in power networks while pursuing
  multiple objectives: balancing production and consumption,
  minimizing energetic losses, keeping people and equipment safe, and particularly avoiding
  catastrophic failures.
  While the importance of the application provides a goal in itself, this challenge also aims
  to push the state-of-the-art in a branch of Artificial Intelligence (AI) called Reinforcement
  Learning~(RL), which offers new possibilities to tackle control problems. In particular,
  various aspects of the combination of Deep Learning and RL called Deep Reinforcement Learning
  remain to be harnessed in this application domain.
  This challenge belongs to a series started in 2019 under the name "\href{https://l2rpn.chalearn.org/competitions}{Learning to run a power network}"~(L2RPN).
  In this new edition, we introduce new more realistic
  scenarios proposed by RTE to reach carbon neutrality by 2050, retiring fossil fuel
  electricity production, increasing proportions of renewable and nuclear energy
  and introducing batteries. Furthermore, we provide a baseline using a state-of-the-art reinforcement learning
  algorithm to stimulate future participants. 
\end{abstract}
~\\

\section{Introduction}
Power Systems enable energy transportation from places where it is produced
(nuclear or fossil power plants, hydroelectric generators, wind turbines, solar panels, etc.)
to places of consumption~(e.g. houses, factories, public lighting, etc.).
It is a vital component of our society; it has become so common that it is often taken for
granted, although it constantly relies on thousands of kilometers of transmission lines and
the work of thousands of people. Power systems are currently facing systemic changes, which bring current
technology to its edge. Recent successes achieved in AI by Deep Learning
techniques~\cite{DBLP:DL}, including Deep Reinforcement Learning~\cite{DBLP:DRL},
have drawn the attention of the Power Systems community for several reasons:
their capacity to learn representations, and their parallelizable architectures.\\
For the fourth edition of the Learning to Run a Power Network challenge (L2RPN'2022), we
look ahead to 2050, in a context of carbon neutrality, by drastically reducing
the share of electricity produced by fossil fuels and increasing the share
of renewable energies in the power system's energy mix.
In this section we briefly review the current context motivating the creation of such
a challenge and the problems posed to the AI community, particularly those resulting from
the massive use of renewable energy.

\subsection{Energy shift}\label{sec:energy_shift}

\paragraph{Global warming}
In the late 2010s, around 85\% of the energy produced came from the combustion of fossil
fuels that emit greenhouse gas such as (but not limited to) $CO_2$.
Those emissions have been consistently growing since the start of the industrial era.
It is nowadays commonly admitted that the negative impact of modern societies on the environment
has become non-negligible since the 1950s. To prevent the irreversible destruction of
an ecosystem, which we need for our survival, it has become urgent to drastically reduce,
among other things, the emission of greenhouse gas~\cite{IPCC-2022}.

\paragraph{Increasingly uncertain power injection patterns}
Political leaders have been pushing toward the development of alternative energy conversion devices
that exploit renewable and low-carbon forms of energy, such as solar radiation and wind. 
Devices that harness those sources of energy have drastically improved over the past two decades,
which has enabled their large-scale deployment. A growing amount of research is dedicated to
investigating the feasibility of a 100\% renewable energy system in the medium term and advocating
for massive use of the latter two technologies.

Unfortunately, solar and wind power come with some drawbacks with regard to their integration
in power networks~\cite{donon:tel-03624628}:

\begin{itemize}
  \item Their production highly depends on the weather.
        This may cause imbalances in the power network.
  \item Our energy storage capacity is low, which promotes
        controllable generators, as opposed to intermittent generators
        such as solar and wind power. It is therefore mandatory to have
        controllable generators in reserve.
\end{itemize}

\subsection{Complexity of power network operations}
The electric power network can be broken down into main functions:
production (power generation), transport (power lines), and consumption (end users).
The transport part is usually split into the "transmission system"
(long distances, e.g. from a power plant to a city)
and the "distribution system" (local scale, e.g. within a city).\\
RTE is in charge of managing the French transmission system in real-time
and ensures that the production equates to the consumption.
It anticipates the impacts of potential outages, whether these are planned or accidental.
Dispatchers (highly trained engineers) ensure the system's security by performing several
actions, including \cite{donnot:tel-02045873}:

\begin{itemize}

  \item managing power overflows (which can endanger trees, roads, infrastructures or passers-by)
        and preventing cascading failures (leading to blackouts of the whole system),
        by changing interconnection patterns of transmission lines, to redirect power flows);
  \item asking producers or consumers to change what they inject into the power network
        (for example, remunerating a producer at a specific localization to avoid a local overload);
  \item in the future, maybe modifying the amount of power produced or absorbed by storage units,
        such as batteries;
  \item when required, limiting the amount of energy injected by renewable generators
        (such as wind or solar) in case of overproduction or local issues for example.
\end{itemize}
In all cases, dispatchers have to rely on their thorough understanding of the system.
Current optimization-based methods are struggling with the complexity of both problems,
and some satisfying heuristics exist or are in the process of being experimented.
The hope is that AI could assist dispatchers in making better decisions to efficiently
control the power network and keep all equipment in security.

\begin{figure*}[t]
  \centering
  \includegraphics[scale=0.25]{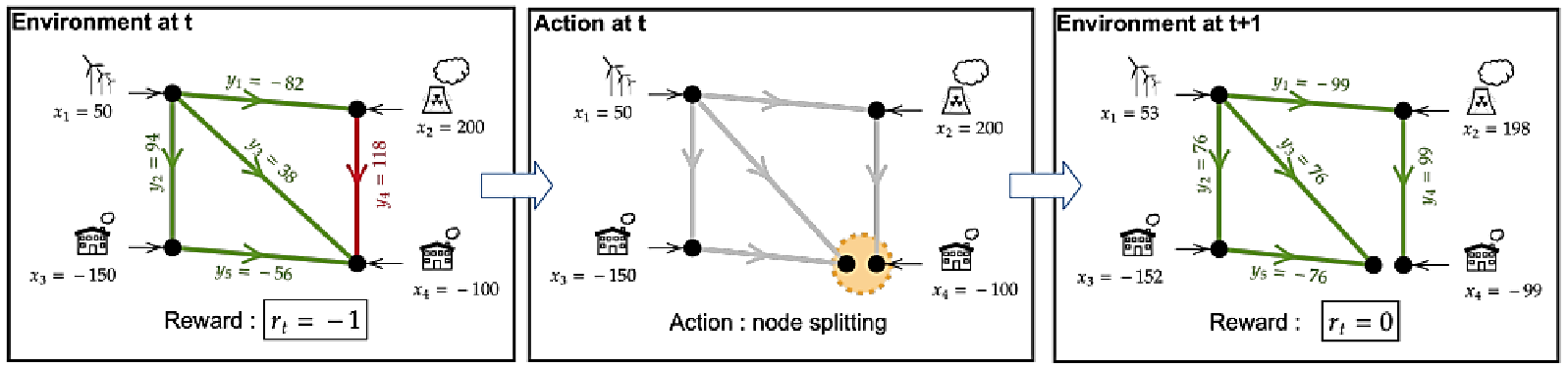}
  \caption{{Power system operation:} The task of dispatchers is to monitor the power network and make eventual changes to ensure safe network operation with no line overflow. If in the environment at time t (left) a line is overflowing (indicated in red), a corrective action may be taken (center), such as a "node splitting", resulting in restored "power network safety" in the environment at time t+1 (right).
           Borrowed from \cite{l2rpn_analysis}.}
  \label{fig:operation}
\end{figure*}

\section{Previous challenges}
The "Learning to run a power network" (\href{https://l2rpn.chalearn.org/}{L2RPN})
challenge \cite{l2rpn_design, l2rpn_analysis} is a series of competitions that model
the sequential decision-making environments of real-time power network operations,
as illustrated in Fig.~\ref{fig:operation}. The participants' algorithms must control a simulated power network, in a
reinforcement learning framework.

\paragraph{Power networks and data}
The physical simulation is based on Python's module \href{https://github.com/rte-france/Grid2Op}{Grid2Op}
which we detail in section \ref{sec:env_chronics}.
Power networks of various sizes and topologies are used across competition rounds.

\paragraph{Results and main outcomes}
In the 2019 edition, the provided power network, a slightly adapted version of the IEEE 14-bus
network, was composed of 20 power lines, 11 loads and 5 generators. The winner team of this edition~\cite{2019_winner}
used the Double Deep Q-Learning algorithm~\cite{ddqn} along with imitation learning to initialize the policy.
In 2020, the power network was much more complex. It was composed of 59 power lines, 37 loads and 22 generators.
There were two competitions that year.
The winner team of the first one~\cite{2020_winner1} was a team of experienced researchers in reinforcement learning,
that have won similar competitions in NeurIPS 2018 and 2019: "Learning to Run" and "Learning to Move"~\cite{learning2move}.
First, they perform an action space reduction to 1000 elements using
simple expert systems and initializes a policy parameterized by a feed-forward neural network with
millions of parameters. Then, they train a policy using evolutionary black-box optimization~\cite{blackbox_winner}.
This functioning, using evolution strategies, can be opposed to most standard RL strategies,
such as Deep Q Network (DQN)~\cite{dqn1, dqn2}, which rely on gradient descent.
The winner team of the second one~\cite{2020_winner2} used a policy neural network to select
the Top-$K$ actions and applied an optimization algorithm to choose the best one.

Overall, this latest competition has shown encouraging results about how artificial intelligence
methods can be successfully applied to the problem of operating power networks.
The Grid2Op platform, as a simulator lowering the barriers to entry in the specific domain
of electricity network management, permitted a research team with no knowledge in this domain
to produce a satisfying model. Solutions could be extended to more complex cases, especially as
the power systems are evolving to meet decarbonization, which leads to increasingly difficult
decision problems for operators.

\section{Why a new competition}\label{sec:new_competition}
The French electricity network management company RTE has recently published the
results of an extensive study outlining various scenarios for tomorrow’s French
power management~\cite{futurs_energetiques}. Due to the ecological concerns,
all scenarios mostly rely on nuclear and renewable energies. The Paris region,
Ile-de-France, being particularly concerned, proposed two milestones~\cite{idf}:
\begin{itemize}
    \item \textbf{By 2030}: Reduce by half the dependence on fossil fuels and
          nuclear power in the Ile-de-France region. This would be achieved
          by both reducing the energy consumption by 20\% and by multiplying
          by 2 the energy production from renewable sources.
    \item \textbf{By 2050}: Moving towards a 100\% renewable energy and
          zero carbon region. This would be achieved by both reducing the
          energy consumption by 40\% and by multiplying by 4 the energy
          production from renewable sources.
\end{itemize}

In this context, it is necessary to re-factor the problem tackled by past
L2RPN challenges \cite{l2rpn_analysis}, mainly by updating the simulator and
the data to represent zero carbon scenarios~\cite{futurs_energetiques}.
As explained in section \ref{sec:energy_shift}, such scenarios are even harder to control,
therefore advancing Artificial Intelligence methods could be particularly useful.

\section{Competition design}
In order to organize the 2022 edition of Learning to Run a Power Network, we need
an environment that will simulate the behavior of a power system during a defined time
(e.g. one week) that we will call scenario. To simulate such a scenario, the environment
needs data and more specifically time series describing the electricity injections in the power network
(referred to as chronics). In this section, we describe the specificities of the simulation environment
used as well as the chronics. In addition, we also look at the details of the organization
of the competition on an online platform and the metric used to rank the participants.

\subsection{Simulation environment and chronics}\label{sec:env_chronics}
\paragraph{Grid2Op}
To run a L2RPN competition, we need a library capable of simulating
a power system in a reinforcement learning framework. RTE has therefore developed
Grid2Op~\cite{grid2op}, a Python module that casts the operational decision process into
a Markov Decision Process $(S, A, P_a, R_a)$~\cite{MDP}.
Grid2Op will therefore discretize the time of a scenario into a list of
states corresponding to a time step of 5 minutes. For example, a one day
scenario will be discretized in $24 * 60 / 5 = 288$ time steps.
Then, for a state $s_t \in S$ and an action $a_t \in A$, Grid2Op will calculate
$s_{t+1}$ i.e. the power flow (the amount of electricity flowing on each power line) at time $t+1$.
For this, it will need the chronics at time $t$.
We detail the generation of these data in the next paragraph.
In addition, Grid2Op uses the Gym interface developed by OpenAI~\cite{gym}
to interact with an agent. Also, a set of startup notebooks is available to facilitate its handling.
Thanks to these two points and as previous editions have confirmed, future participants don't
need to have strong knowledge in the field of power systems to create
efficient agents. This corresponds perfectly to our desire to create a competition open to all and
oriented towards reinforcement learning.

\begin{figure}[b]
  \centering
\vspace{-0.5cm}
  \includegraphics[scale=0.42]{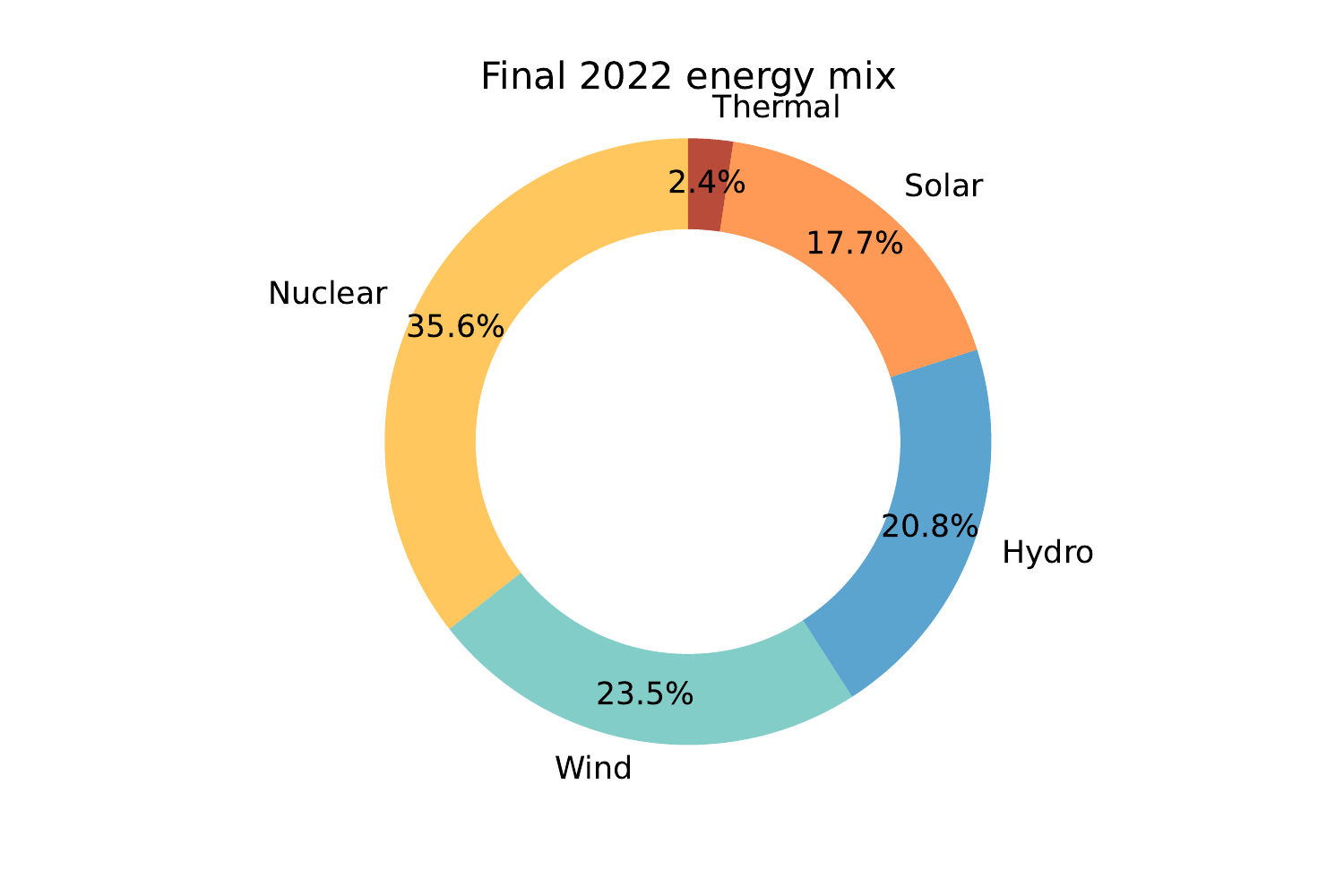}
  \vspace{-0.5cm}
  \caption{L2RPN 2022 energy mix over a year.}
  \label{fig:2022_em}
\end{figure}

\begin{figure*}
  \centering
  \includegraphics[scale=0.37]{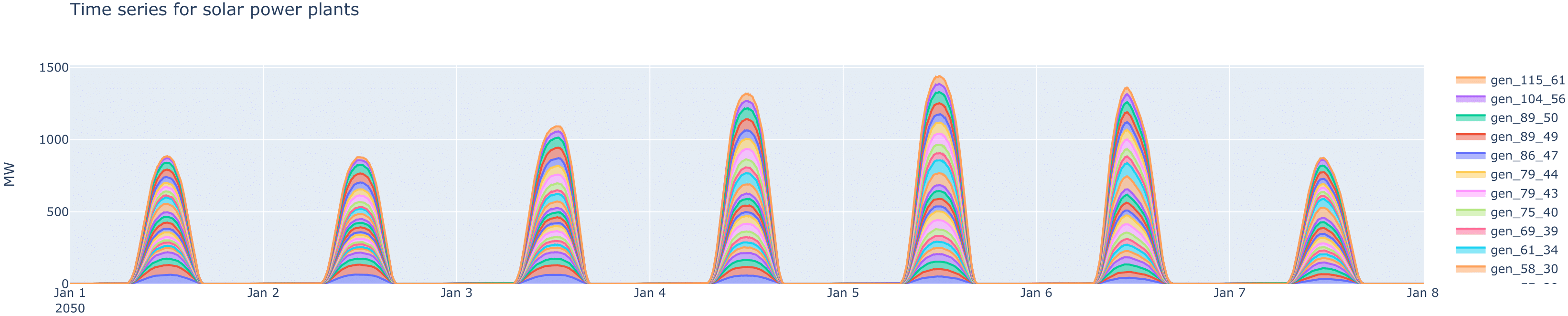}
  \caption{Example of time series representing the energy produced by each solar power plant at each time step.}
  \label{fig:time_series}
\end{figure*}

\paragraph{Chronics}
As described in the previous paragraph, to make Grid2Op work,
we need to generate time series describing the electricity injections into the power network.
These time series are referred to as chronics.
An injection is the amount of electricity that is injected into the power network
by generators, loads and batteries.
The generators inject a positive amount of electricity while the loads inject a negative amount.
Batteries can inject either a negative or positive amount of electricity
depending on whether they are storing or delivering electricity. The sums
of the injections must be equal to 0 at all times for the power network to work
(taking into account the loss of electricity in heat).
To generate these chronics, we need data concerning
the architecture of the power network, the weather,
the consumptions and the generators
(e.g. their types and maximum production).
These data, especially about the consumptions and the weather, come from RTE studies.
Concerning the power network, we are using
an even more complex power network than in previous years.
It is composed of 186 power lines, 91 loads and 62 generators. In addition, we have added
7 batteries that can be used by agents to store and deliver electricity.
Then, a library created by RTE named Chronix2Grid~\cite{chronix2grid},
uses these data to generate chronics.
Fig~\ref{fig:time_series} is a example of such chronics.
From the chronics, we deduce the energy mix of our power system:
\begin{equation}
  em_{gt} = \frac{\sum_{g \in G_{gt}} \sum_{t=0}^T \epsilon_{g, t}}{\sum_{g \in G} \sum_{t=0}^T \epsilon_{g, t}} ~,
\end{equation}
\noindent
where $gt$ is a generator type that belongs to the set $\{nuclear, solar, wind, thermal, hydro\}$,
$em_{gt}$ is the percentage of electricity produced by generators of type $gt$,
$G_{gt}$ is the set of the generators of type $gt$, $G$ is the set of all the generators, $T$ is
the maximum time step of the scenario, and $\epsilon_{g, t}$ is the injection of generator $g$ at time
step $t$.\\
To generate this edition's chronics, we privileged
renewable generators and penalized the use of
fossil fuel generators (referred to as \textit{thermal}) in Chronix2Grid.
In addition, we have given it the possibility to curtail the excess
of renewable energy, if needed, when creating the chronics.
This has allowed us to significantly increase
the power of renewable generators. As a result, we were able to generate chronics
with an almost carbon-free energy mix as shown in Fig.~\ref{fig:2022_em}.
With less than 3\% of electricity generated by fossil fuels,
this energy mix is very satisfactory for our competition, so we have generated
32 years of scenarios that are available to participants to train their agents.
Moreover, they can generate more scenarios with the same
specifications through Grid2Op.

\subsection{Hosting on Codalab}
To facilitate participation, we implemented the L2RPN'2022 competition on \href{https://codalab.lisn.upsaclay.fr/}{Codalab} as follows:
\begin{itemize}
  \item {\bf Competition with code submission}: RL agents capable of controlling the power network will be blind
        tested on the platform with new scenarios not known to the participants. These scenarios were
        carefully chosen to be representative of the different problems encountered by power network operators.
  \item {\bf Starting kit}: We provide a set of tools and tutorials to help participants
        getting started, including power network visualization and diagnosis tools
        (\href{https://github.com/rte-france/grid2viz}{Grid2Viz}),
        and a white paper describing the problem and baseline methods.
        A sample submission with the code of a baseline agent, following a designated API, is provided.
        Sample scenarios are supplied. They are chosen with the same criteria as those used to test the agents
        on Codalab, but they are not the same. The starting kit is available
        \href{https://codalab.lisn.upsaclay.fr/my/datasets/download/4192412f-4f83-49ff-80d7-595620534410}{here}.
  The execution of a task consists in repeating the following RL-style steps, until time is out or a blackout occurs:
  \begin{itemize}
      \item[-] suggestion = agent.act(observation)
      \item[-] observation = environment.step(suggestion)
  \end{itemize}
  \item {\bf Protocol}: Participants will need to train their agent on their own machine
        or cloud server, and submit trained agents.
        We will use a three-phase competition protocol: (0) warm-up phase: participants try
        the starting kit, can ask for modifications of the computational resources and
        the available packages; (1) development phase: packages and computational
        resources are frozen; participants get feedback on their submissions on a leaderboard,
        and (2) final phase: a single final submission is evaluated on new unseen scenarios.
        This evaluation is the only one counting for the final ranking. 
  \item  {\bf Timeline}: The project was accepted as an official IJCNN/WCCI'22 in January 2022. We opened the warm-up phase on June 15, and the development phase July 5. The final test phase will start September 15, and the results will be revealed September 30.
\end{itemize}

\subsection{Metric}
\label{sec:metric22}
To rank the participants, we need a {\bf score function}
which assigns a real number to each agent evaluating its performance.
To that end, we created a score function that is the average of these three cost functions over the
test scenarios:

\begin{itemize}
  \item \textbf{Cost of energy losses}: Calculated by multiplying the amount
        of electricity lost due to the Joule effect by the current price of the MWh.
  \item \textbf{Cost of operation}: Sum of the costs of the agent's
        actions.
        Operations involving changes in the production of electricity
        have a cost that depends on the energy market.
        The use of batteries has a fixed cost per MWh.
  \item \textbf{Cost of blackout}: If the agent did not manage the power network
        until the end of the scenario, this cost is calculated
        by multiplying the amount of electricity left to supply by the current price of MWh.
\end{itemize}

Note that, as expected, the cost of a blackout is much higher than the two other costs,
which means that an agent who succeeds in a scenario will always have a better score than an agent who has
not succeeded, even though its actions are less costly.\\
Moreover, our score function is normalized so that it is to be maximized and is
between the bounds $[-100, 100]$. A score of $0$ corresponds to an agent that does nothing
at each time step. Having a positive score is already pretty good.

\subsection{Setting recap}\label{sec:recap}
We summarize the setting of the optimization problem to be solved, at every step:

\begin{itemize}
    \item \textbf{Observation space}. Complete state
          of the power network: all information over power nodes
          (electricity produced and consumed), flows of each power lines,
          and more. 
    \item \textbf{Action space}. Four types of actions allowed: 
    \begin{align}
1.~ & \text{Line status (line connection/disconnection).} \nonumber \\
2.~ & \text{Topology changes (node splitting).} \label{eq:actions} \\
3.~ & \text{Power production changes/curtailment (of generators).} \nonumber \\
4.~ & \text{Storage changes (storage or delivery from batteries).} \nonumber 
\end{align}
          For the 2022 edition, the action space still contains over 70,000 discrete actions
          (topology changes) and 69-dimensional continuous action space (production changes).
    \item \textbf{Reward}. The participants are free to design their own reward function.
          However, the leaderboard metric is defined in Section \ref{sec:metric22}.
    \item \textbf{Game over condition}. A game over is triggered if total demand is not met anymore
    (taken into account in the metric as "cost of blackout").
\end{itemize}

\section{Baseline}\label{sec:baseline}
In addition to providing a more complex power network
with a carbon neutral energy mix, for the 2022 edition of L2RPN
we also provide a baseline using reinforcement learning (RL).
It has a relatively simple architecture but performs quite well on the validation scenarios.
The goal is on one hand, to give an example of a simple agent using
reinforcement learning and, on the other hand, to stimulate the competition.
Furthermore we also wanted to give a working example to the participants to
leverage the new types of actions at their disposal: curtailment and action
on storage units. This is why our baseline agent only uses these new actions.

\subsection{Prior art}
The efficiency of reinforcement learning (RL) to solve complex sequential
decision problems has been demonstrated many times.
For example, Go~\cite{master_go}, Chess~\cite{master_chess}
and even complex video games like Dota 2~\cite{master_dota}
have recently made great strides, thanks to RL algorithms.
The use of RL in power system operation
has also been illustrated~\cite{ernst_2004, zhang_2020}, and the winners of previous Learning to Run a Power Network competitions~\cite{2019_winner, 2020_winner1, 2020_winner2}
have used RL approaches.

\subsection{Architecture}
\label{sec:archi}
\begin{figure}[b]
  \vspace{-0.5cm} 
  \hspace*{-2.5cm}
  \includegraphics[width=1.5\columnwidth]{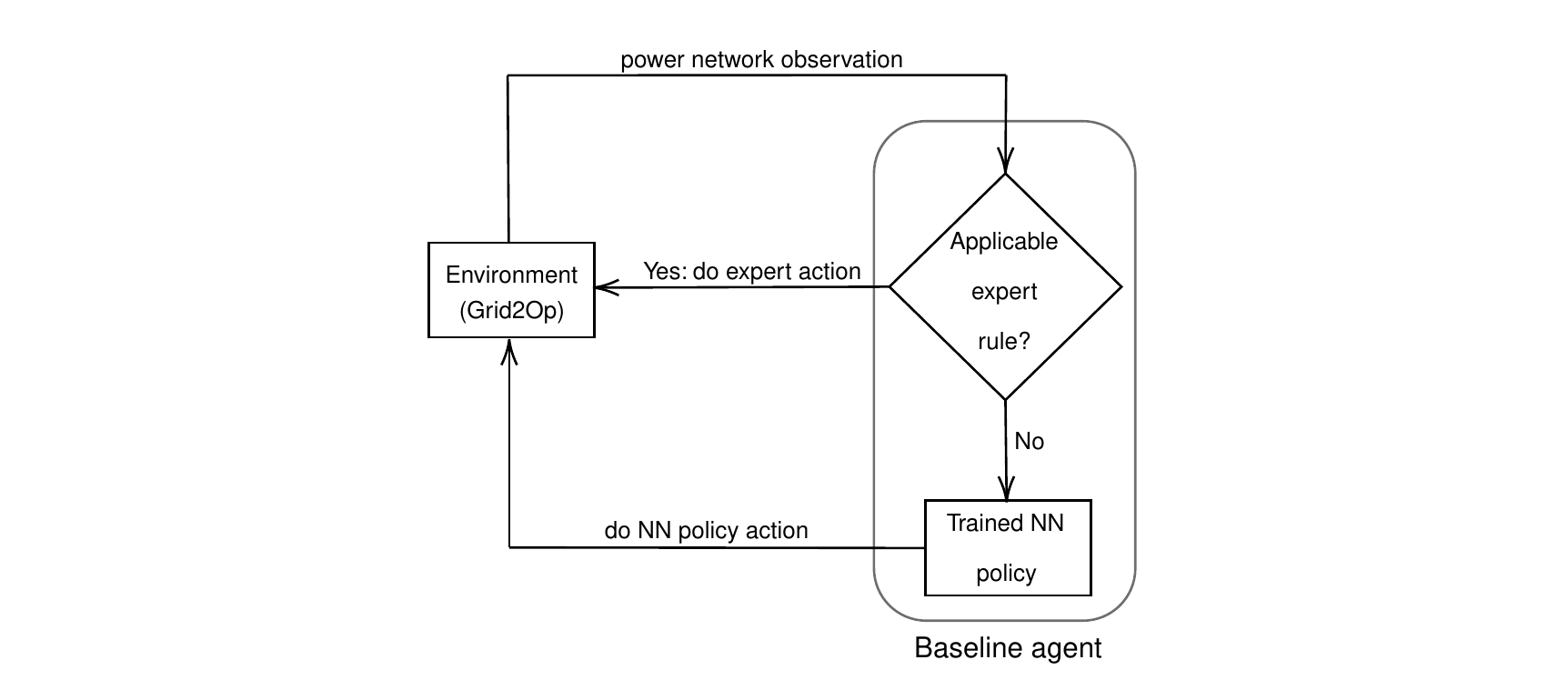}
  \caption{Overview of our baseline architecture}
  \vspace{-1cm} 
  \label{fig:agent_archi}
\end{figure}

The architecture of our baseline agent is illustrated in Fig.~\ref{fig:agent_archi}.
It takes all data concerning power lines, generators, and storage units from the
power network state as ``observation'', and first checks
whether it improves the system state by performing an ``obvious
action'' with ``expert rules'' (specifically: line re-connections or do-nothing).
If not fruitful, the agent uses a ``trained policy'' to choose another action, involving a parameterized neural network, trained with an Actor-Critic algorithm~\cite{actor-critic}.
This architecture is more efficient than just using a trained policy network
because simple expert rules
maintain the power network well in most situations.

\subsection{Proximal Policy Optimization}
The RL part of the agent
focuses on continuous actions from Eq.\ref{eq:actions}, which are: 3. curtailment
and  4. storage units, described in more detail Sec.~\ref{sec:recap}.
We chose the Proximal Policy Optimization (PPO) algorithm~\cite{ppo},
which has had success in previous editions of our competition, and is generally
known to be very efficient in complex environments with a continuous action space.
It is used e.g. in MuJoCo~\cite{mujoco} or Roboschool~\cite{ppo}.\\
PPO is a policy gradient and Actor-Critic algorithm: it uses an approximate value
of the cumulative sum of the rewards (say $V_t$) to "criticize" and update the policy.
Generally, this type of algorithm has two neural networks,
one which returns $V_t$, called \textit{critic network}, and one which
delivers the policy, called \textit{policy network}.

In the vanilla policy gradient algorithm,
the $V_t$ value is used to compute an estimator of the quality of the action
chosen by the policy at time $t$:
\begin{equation}
  \widehat{A}_t = \sum_{k=0}^{T-t} \gamma^k r_{t+k} - V_t
\end{equation}
where $T$ is the number of time steps of the episode, $0 \leq \gamma \leq 1$ is the
discount parameter, and $r_t$ is the reward obtained at time $t$.
$\widehat{A}_t > 0$ means that the \textit{critic network} predicts that
the action chosen by the policy at time $t$ is good so the algorithm
will update the policy in order to increase the probability of doing this action
and vice-versa if $\widehat{A}_t < 0$.
The \textit{critic network} is used only during training
and it is trained using the MSE loss function to approximate
the cumulative sum of rewards.
However, at the beginning, because of the random initialization
of the weights of the \textit{critic network}, the value of $\widehat{A}_t$ is random.
Thus, many times, the algorithm will update
the policy in the wrong direction, which explains the instability and the slow convergence
of the vanilla policy gradient algorithm.

PPO tries to solve this problem by proposing an objective that optimizes
the policy, while penalizing too large updates.
The objective is to find the $\theta$ parameters that maximizes:
\begin{equation}
    \mathc{L} = \mathbb{E}_t \left[ min \left(\widehat{A}_t q_t(\theta),
                                          \widehat{A}_t clip \left(q_t(\theta), 1-\epsilon, 1+\epsilon  \right) \right) \right]
    \label{eq:l_ppo}
\end{equation}
Where $\epsilon$ is a hyperparameter ($0.2$ in the original PPO paper) and
$q_t(\theta)$ is the ratio of the probability of doing the
action $a_t$ at state $s_t$ between the new policy
parameters $\theta$ and the previous $\theta_{old}$:
\begin{equation}
    q_t(\theta) = \frac{\pi_\theta(a_t|s_t)}{\pi_{\theta_{old}}(a_t|s_t)}
\end{equation}
Where $\pi_\theta(a_t|s_t)$ is the probability of doing the action $a_t$
at state $s_t$ using the distribution parameterized by $\theta$.
When $q_t(\theta) > 1$ it means that the probability of doing the action
$a_t$ at state $s_t$ became more probable with the new distribution parameters.
In the other hand, $q_t(\theta) < 1$ means that the probability of doing the action
$a_t$ at state $s_t$ became less probable.\\
Thanks to this term and the $min$ and $clip$ functions,
PPO limits the chance of making destructively large policy updates
to prevent the agent from going in a direction that looks good
but turns out to be a bad one. This behavior is very well explained
by Fig. 1 of the original paper~\cite{ppo}.
PPO is therefore more stable and trains faster than
vanilla policy gradient algorithm.

\subsection{Experimental setting}\label{sec:exp_settings}
\paragraph{Actions}
Four types of actions are possible in principle (Equation\ref{eq:actions}), but our baseline method excludes actions of type 2 (node splitting).
The rule-based part of the agent performs type 1 actions (line reconnections);
the RL part of our baseline agent
 only performs two types of actions: 3.~curtailment actions and 4.~storage actions.
The former can modify the production of renewable generators and the latter
can store or deliver electricity to the batteries. Both actions are continuous.

\begin{figure}
    \centering
    \vspace{-0.5cm}
    \includegraphics[scale=0.423]{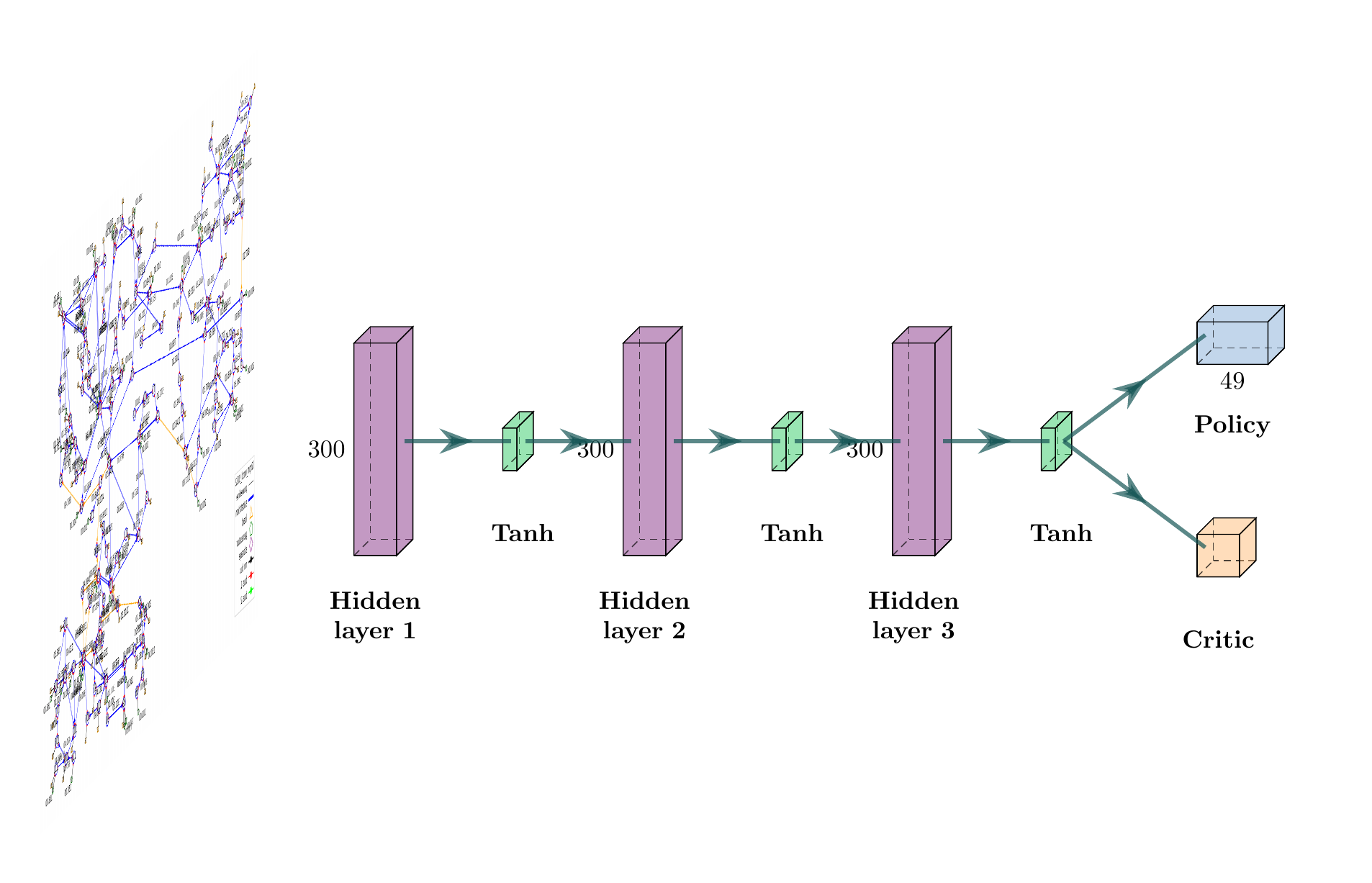}
    \vspace{-0.8cm}
    \caption{Architecture of our baseline agent Actor-Critic network}
    \label{fig:nn_archi}
\end{figure}

\paragraph{Actor-Critic neural network}
Our neural network is a Multi-Layer-Perceptron.
Its architecture is illustrated by Fig.~\ref{fig:nn_archi}.
It is composed of 3 hidden layers of 300 neurons each which are shared
by the Critic network and the Actor network. The input shape is $1225$,
the output shape of the Critic network is $1$ and the output shape of the
policy network is $49$, which is consistent with the possible actions
detailed above since there are $42$ renewable generators and $7$ batteries.
We use the $tanh$ activation function between the hidden layers.

\paragraph{Reward}
The reward used to train our PPO agent is defined as:
if the game is over, it returns the ratio between number
of time steps survived by the agent and number of time steps of the scenario;
otherwise, it returns 0.

\paragraph{Expert rules and associated hyperparameters (HP)}
The expert rules of our baseline agent shown in Fig.~\ref{fig:agent_archi} are:
\begin{itemize}
    \item[-] {\bf Reconnect all possible power lines.} Some lines may have been disconnected because of a
          maintenance for example. We assume that the more lines are connected, the less likely
          it is that the power network will be overloaded.
          
    \item[-] {\bf Do nothing else (if possible); HP \textit{safe\_max\_rho}.} If the power network is not close to being overloaded by doing nothing, then perform the \textit{do nothing} action.
          Let $l$ be the most loaded power line and $\rho_l$ the value of its load (ratio between amount of electricity passing through $l$
          and its maximum capacity). 
          If $\rho_l <$ \textit{safe\_max\_rho}, \textit{do nothing}, else use the PPO policy.
    \item[-] {\bf Limit action impact;  HP \textit{limit\_cs\_margin}.} The HP \textit{limit\_cs\_margin} limits the impact of an action on the power network. This last rule is motivated by empirical observations that RL agents have trouble producing smooth actions and change multiple generator states in a single action.  This often leads to a premature ``game over" by violating some constraints of the environment. 
\end{itemize}

\paragraph{Neural Network software and hyperparamaters} To implement our PPO agent we used the Python Stable Baselines 3
library~\cite{stable-baselines3}.\footnote{version 1.5.0.
The code to reproduce these results and
Fig.~\ref{fig:score_rho}, \ref{fig:score_cs_margin}
and \ref{fig:score_set} is available
\href{https://github.com/gaetanserre/L2RPN-2022\_PPO-Baseline}{here}.}
The training hyperparameters are described in Table~\ref{tab:hyperparameters}\footnote{
oracle: for training, we rely on the environment to limit
the impact of the action in case the action is too heavy.
It is not possible at inference time as the environment
cannot be modified by our agent.}.\\
\begin{table}[b]
    \centering
    \caption{Baseline agent training hyperparameters}
    \label{tab:hyperparameters}
    \begin{tabular}{c c}
    \hline
        Model type & Multi-Layer-Perceptron\\
        \hline
        Input shape & 1225\\
        Output shape & (1, 49)\\
        Shared hidden layers & 3 of 300 neurons each\\
        Loss function & PPO loss for policy and MSE for critic\\
        Batch size & 16\\
        Environment steps & 16\\
        Gamma & 0.999\\
        Epochs loss optimization & 10\\
        Optimizer & ADAM\\
        Learning rate & $3e^{-6}$\\
        \textit{safe\_max\_rho} & 0.2\\
        \textit{limit\_cs\_margin} & oracle\\
    \hline
  \end{tabular}
\end{table}

\paragraph{Training data} Preliminary experiments that we conducted revealed that training on all scenarios (available from the public dataset of the competition) resulted in worse results than ``cherry picking'' scenarios. To alleviate this problem, we limited training to scenarios taken from the most
difficult week of the year (one week in February, when power consumption is high).
More systematic experiments to optimize the training curriculum are under way.

\paragraph{Validation \& Test data}  The results presented in this paper use the validation set of the ``development phase'' to select hyperparameters (of the expert rules and the neural network) and the test set of the "test phase" of
our competition on Codalab\footnote{\href{https://codalab.lisn.upsaclay.fr/competitions/5410}
{https://codalab.lisn.upsaclay.fr/competitions/5410}} to report results.

\subsection{Results}\label{sec:results}
We trained 14 agents for 10 million iterations.
In the following figures, we compare several agents with
the \textit{Do Nothing} agent (does nothing at
each time step) and the \textit{Expert only} agent
(uses only our experts rules defined in Sec.~\ref{sec:exp_settings}).

\begin{figure}
    \centering
    \includegraphics[scale=0.36]{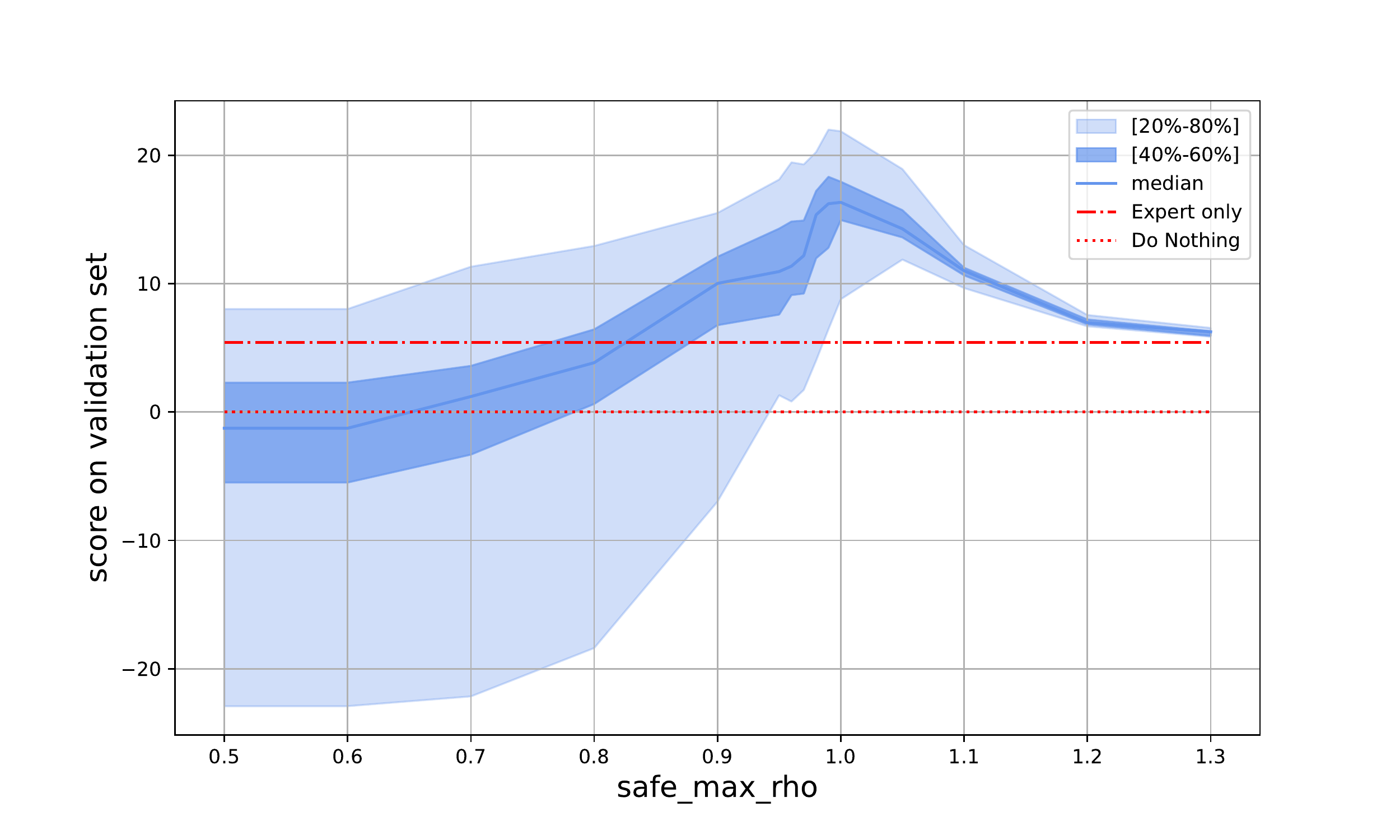}
    \caption{Baseline score
            (as defined in Sec.~\ref{sec:metric22})
            averaged over all validation scenarios
            as a function of \textit{safe\_max\_rho}
            at evaluation time.
            $limit\_cs\_margin = 60$.}
    \label{fig:score_rho}
\end{figure}

\begin{figure}
    \centering
    \includegraphics[scale=0.36]{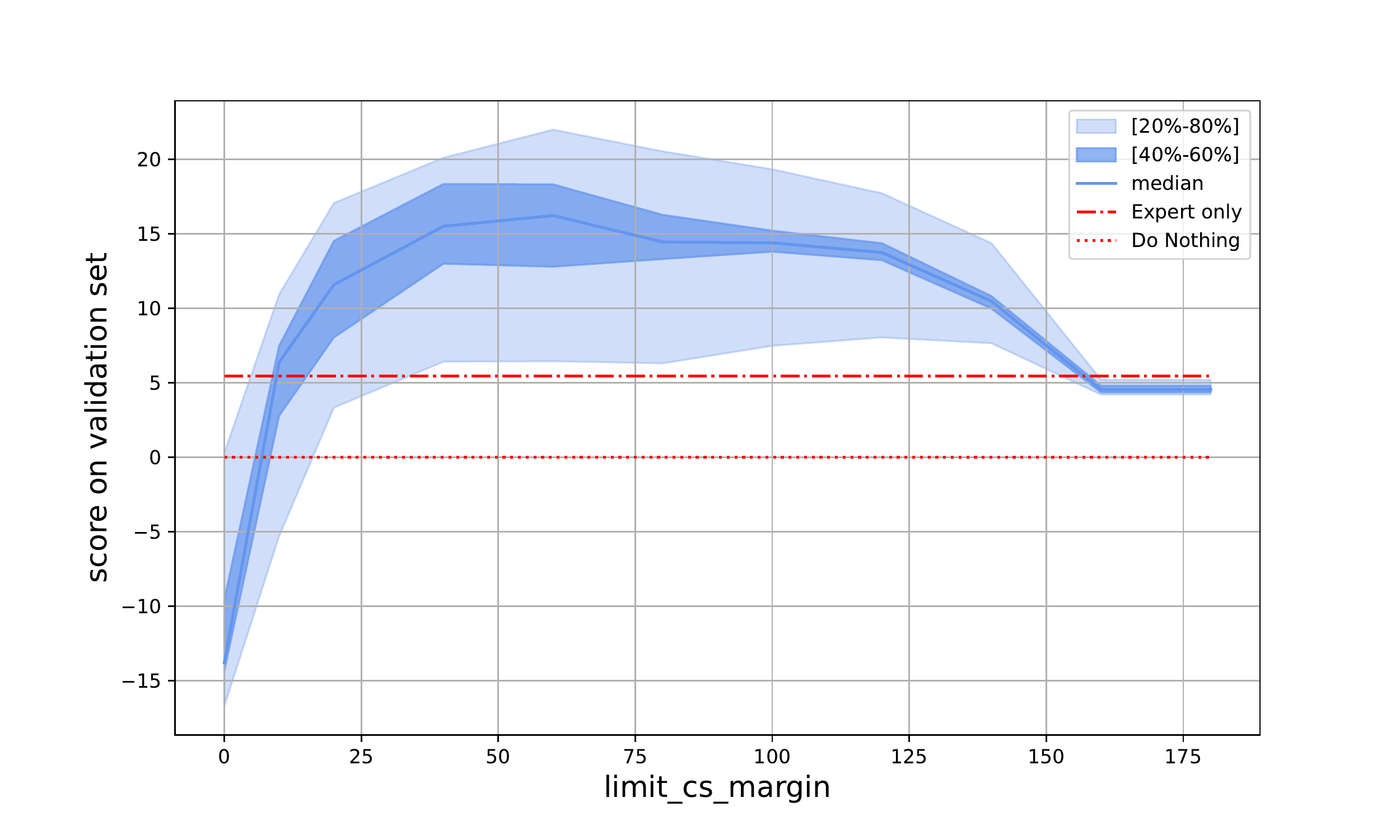}
    \caption{Baseline score
            (as defined in Sec.~\ref{sec:metric22})
            averaged over all validation scenarios
            as a function of \textit{limit\_cs\_margin}
            at evaluation time.
            $safe\_max\_rho = 0.99$.}
    \label{fig:score_cs_margin}
\end{figure}

To improve the score of our agent (Fig. \ref{fig:agent_archi}), we tuned (using validation data) the two hyperparameters described in Sec.~\ref{sec:exp_settings}: \textit{safe\_max\_rho}
and \textit{limit\_cs\_margin}. They limit the impact of the action of our RL agent on the power network, using expert rules, to avoid erratic actions triggering early "game over".\\
To find the best combination of these hyperparameters,
we set
$limit\_cs\_margin = 60$ (a medium value chosen empirically) and look for the best
corresponding value of \textit{safe\_max\_rho}.
As illustrated in Fig.~\ref{fig:score_rho},
this value is $0.99$.
We then set $safe\_max\_rho = 0.99$
in order to look for the best
corresponding value of
\textit{limit\_cs\_margin}.
As illustrated in Fig.~\ref{fig:score_cs_margin},
this value is $60$.
With this method, the best combination of
these hyperparameters is $safe\_max\_rho=0.99$ and
$limit\_cs\_margin=60$.
A large \textit{safe\_max\_rho} results in using
the ``RL part of our agent'' only in critical states
of the power network, thus avoids doing potentially
destructive actions, when doing nothing is enough.
In addition, 60 is an intermediate value of
\textit{limit\_cs\_margin}, which is a good
compromise between limiting and preserving the action.
With these parameters, our best agent has a score of 22.46
on the validation scenarios and 26.80 on the scenarios
of the test phase of our competition.
This agent is thus much better than the \textit{Do Nothing}
or \textit{Expert only} agents
(higher scores are better).\\

\begin{figure*}
  \centering
  \includegraphics[scale=0.24]{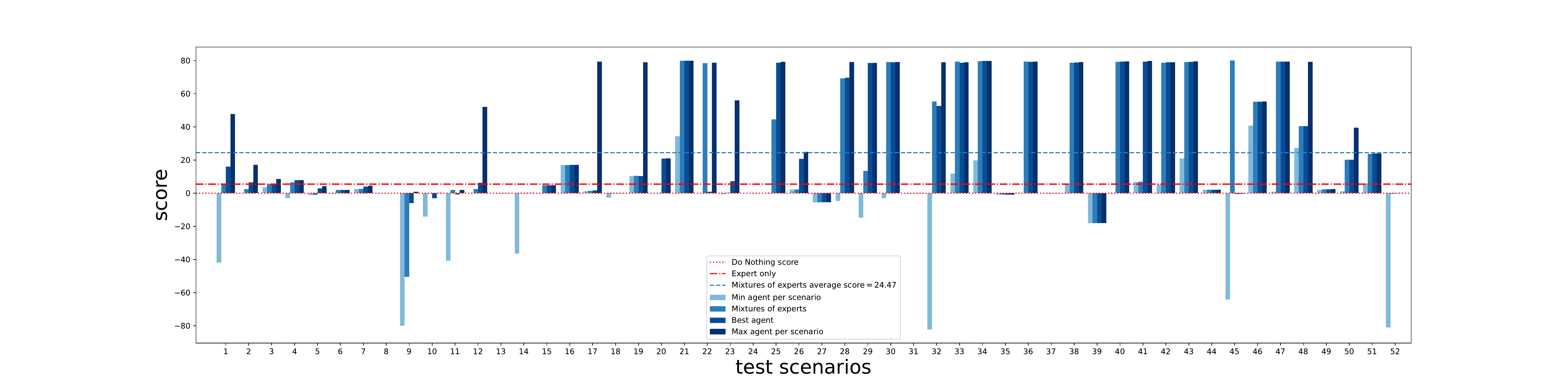}
  \caption{Score of our "mixture of expert agents" algorithm
          over all validation scenarios compared with
          the best and worst agent in the set
          at each scenario and our best baseline agent.}
  \label{fig:score_set}
\end{figure*}

To take advantage of the stochasticity of the score
of our agent, depending on its initialization
and evaluation scenarios,
we created a mixture of experts to choose the best action, taking advantage of the strengths of various RL agents.
Grid2Op allows us to estimate of the reward
obtained if we do the action $a$ at state $s_t$.
Our mixture of expert algorithm therefore implements a "look ahead" policy that simulates
the actions of the available RL agents to choose the
action that brings the best approximation of the reward.
To evaluate this strategy, we have used 14 instances of our baseline trained previously.
Fig.~\ref{fig:score_set} illustrates the performance of our mixture of experts algorithm on the
validation set, compared with other agents.
This algorithm obtains a score of 23.58 on the validation
set and 24.47 on the scenarios
of the test phase of our competition.
These scores are quite similar to those of our baseline agent.
However, in some scenarios, our mixture of agents
outperforms any instance of our baseline.
This strategy looks promising and many enhancements are possible.

\section{Conclusion}
In this paper, we presented the design of the fourth edition of
"Learning to Run a Power Network challenge", focusing on
"energies of the future and carbon neutrality".
This competition targets the real-world problem
of ensuring the safety of power networks, using a lot of renewable
energies and several batteries, with a focus on real-time operations.
We provided a baseline agent to the participants,
combining heuristic rules and a trained RL agent,
to lower the barrier of entry and stimulate participation.
This baseline performs quite well, but could still be improved. 
Indeed, we show in this paper a first promising avenue: creating
random mixtures of RL agents. This could be further enhanced by
specializing the RL agents on subsets of scenarios, e.g.,
around given times of the year. Other improvements could be made
by exploiting actions of type 2: Node splitting.
Finally, more sophisticated but slower optimization algorithms
could be used off-platform to initialize various specialized policies.
On our side, we are conducting more experiments
to understand why agents benefit from training on well-chosen
scenarios, and whether performing a kind of "curriculum learning",
with progressively increasing difficulty in scenarios, may help.
However, our role as organizers is to bootstrap the competition with
a reasonably good agent, but leave room for improvement.
Hence, we did not provide our latest and greatest agent.
The final results of the challenge and post-challenge analyses
will be included in this paper at revision time. The challenge
platform will remain open beyond the termination of the challenge
as an ever lasting benchmark, and we hope to continue organizing
challenges in this series with the feed-back of participants
and the power network community.

\subsubsection*{Aknowledgements} 
We are grateful to Alessandro Leite, Farid Najar, and Sébastien Treguer for stimulating discussions. This project is co-organized by RTE France and Université Paris-Saclay, with support of
  Région Ile-de-France, TAILOR EU Horizon 2020 grant 952215, ANR Chair of Artificial Intelligence HUMANIA ANR-19-CHIA-0022, and ChaLearn. 

\bibliographystyle{plain}
\bibliography{refs}

\end{document}